\documentclass[10pt,twocolumn,letterpaper]{article}
\usepackage[accsupp]{axessibility}
\usepackage{iccv}              

\newcommand{\todo}[1]{{\color{red}#1}}

\definecolor{cvprblue}{rgb}{0.21,0.49,0.74}
\definecolor{iccvblue}{rgb}{0.21,0.49,0.74}
\definecolor{ouryellow}{rgb}{0.84, 0.62, 0}
\definecolor{ourgreen}{rgb}{0.05, 0.55, 0.26}
\definecolor{ourred}{rgb}{1, 0, 0}
\definecolor{scamgreen}{rgb}{0.84, 0.91, 0.84}
\definecolor{scampurple}{rgb}{0.88, 0.84, 0.91}
\usepackage[pagebackref,breaklinks,colorlinks,allcolors=iccvblue]{hyperref}
\usepackage{multirow}
\usepackage{multicol}
\usepackage{makecell}
\usepackage{graphicx}
\usepackage{xcolor}
\usepackage{threeparttable}
\usepackage{enumitem}
\usepackage{colortbl}
\usepackage[para]{footmisc}
\renewcommand{\todo}{\textcolor{black}}

\def\paperID{7017} 
\def\confName{ICCV}
\def\confYear{2025}
\title{Closed-Loop Transfer for Weakly-supervised Affordance Grounding}

\author{
  \textbf{Jiajin Tang}$^1$$^{*}$, 
  \textbf{Zhengxuan Wei}$^1$$^{*}$, 
  \textbf{Ge Zheng}$^1$,
  \textbf{Sibei Yang}$^2$$^{\dagger}$\\
\textsuperscript{1}ShanghaiTech University \\
\textsuperscript{2}School of Computer Science and Engineering, Sun Yat-sen University \\
\texttt{\normalsize{\{tangjj, weizhx2022\}@shanghaitech.edu.cn}}\hspace{0.5cm}\texttt{\normalsize{yangsb3@mail.sysu.edu.cn}}\\
}

\begin{document}
\renewcommand{\thefootnote}{\fnsymbol{footnote}}
\twocolumn[{
  \maketitle
  \vspace{-20pt}
  \captionsetup{type=figure}
  \centering
  \includegraphics[width=1\textwidth]{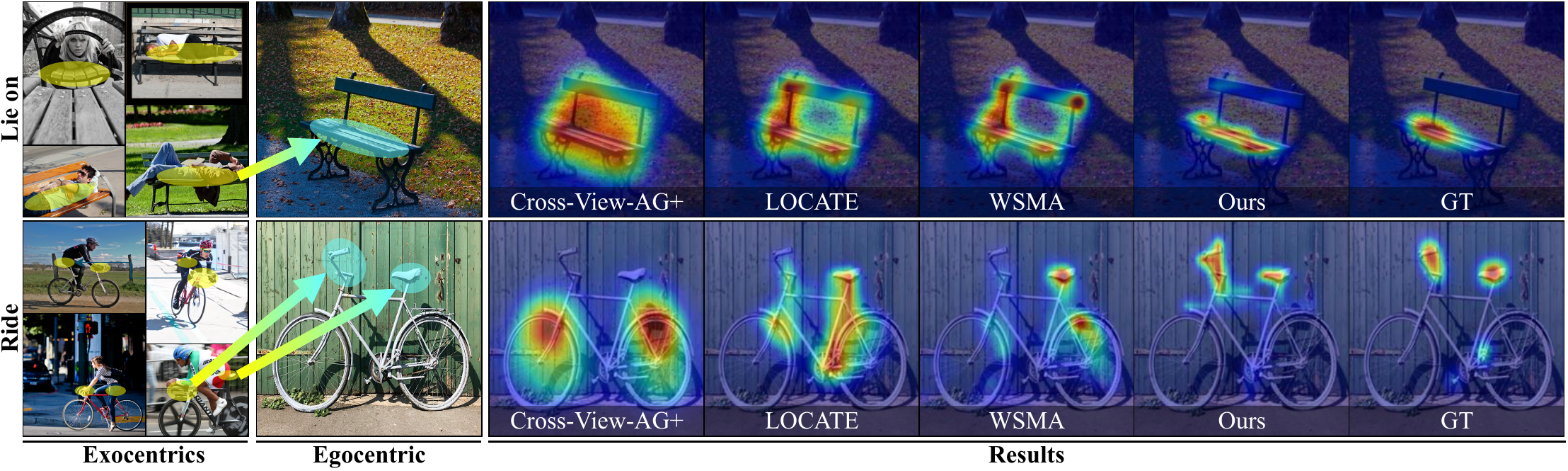}
  \vspace{-20pt}
  \captionof{figure}{Visualization samples for \textit{lie on} and \textit{ride} affordances on egocentric images, comparing state-of-the-art methods with ours.}
  \vspace{15pt}
  \label{fig:1}
}]

\setcounter{footnote}{0}  
\footnotetext{
  $^{*}$ Equal contribution. \quad
  $^{\dagger}$ Corresponding author is Sibei Yang.
}

\begin{abstract}

\begin{figure*}[t]
    \centering
    \includegraphics[width=0.92\linewidth]{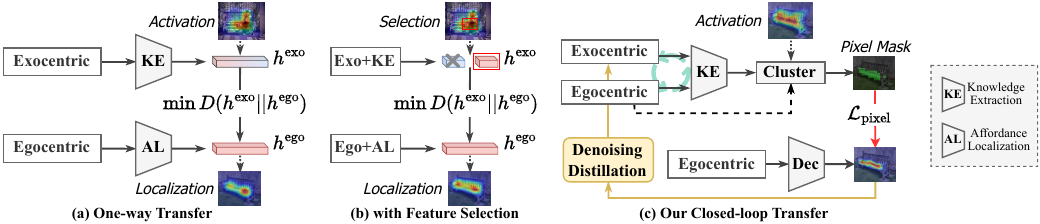}
    \vspace{-6pt}
    \caption{Comparison of (a) one-way exo-to-ego transfer framework~\cite{luo2022learning, xu2024weakly}, (b) one-way transfer with feature selection~\cite{li2023locate}, and (c) our closed-loop transfer framework, LoopTrans. KE refers to extracting human-object interaction knowledge from exocentric images, while AL refers to localizing affordance regions of objects in egocentric images.}
    
    \label{fig:2}
    \vspace{-16pt}
\end{figure*}
\vspace{-3mm}

Humans can perform previously unexperienced interactions with novel objects simply by observing others engage with them. Weakly-supervised affordance grounding mimics this process by learning to locate object regions that enable actions on egocentric images, using exocentric interaction images with image-level annotations. However, extracting affordance knowledge solely from exocentric images and transferring it one-way to egocentric images limits the applicability of previous works in complex interaction scenarios. Instead, this study introduces LoopTrans, a novel closed-loop framework that not only transfers knowledge from exocentric to egocentric but also transfers back to enhance exocentric knowledge extraction. Within LoopTrans, several innovative mechanisms are introduced, including unified cross-modal localization and denoising knowledge distillation, to bridge domain gaps between object-centered egocentric and interaction-centered exocentric images while enhancing knowledge transfer. Experiments show that LoopTrans achieves consistent improvements across all metrics on image and video benchmarks, even handling challenging scenarios where object interaction regions are fully occluded by the human body. Code is available at \small \url{https://github.com/nagara214/LoopTrans}.
\end{abstract}
\vspace{-5mm}    
\section{Introduction}
\label{sec:introduction}
\vspace{-2mm}
The term ``affordance," first introduced by J.Gibson~\cite{gibson1979ecological}, is later formalized in computer vision and robotics to typically describe the ``action possibilities" offered by objects~\cite{burke2010neural,zhu2014reasoning,montesano2008learning,osiurak2017affordance}, such as a knife affords cutting or a bicycle affords riding. Affordance grounding~\cite{chuang2018learning,do2018affordancenet,fang2018demo2vec,myers2015affordance,nguyen2017object,tang2023cotdet} further refines this by not only predicting the actions an object can afford but also pinpointing the specific regions that enable those actions, \eg, a bicycle's handlebars for pushing and both its handlebars and seat for riding (see Fig~\ref{fig:1}). In contrast to most vision perception systems~\cite{zhu2025rethinking,chen2024survey12,huang2023free38,tang2023contrastive70,dai2024curriculum20} that primarily focus on how objects appear~\cite{zheng2023ddcot88,shi2024part2object64,tang2023temporal71}, such as instance and part segmentation, affordance grounding~
emphasizes how objects function. This is essential for embodied intelligent agents to actively interact with and use objects in the real world~\cite{gupta2021embodied,savva2019habitat,franklin1997autonomous,xia2018gibson,bailenson2005independent}, while also facilitating downstream tasks such as object manipulation~\cite{he2017mask,carion2020end} and human-object interaction~\cite{gkioxari2018detecting,chao2018learning}. 

Humans can infer precise affordance grounding~
on objects across diverse actions and environments, even performing unfamiliar interactions with novel objects, simply by observing others interact with them. To mimic this learning process, we follow a practical weakly-supervised affordance grounding~
setting~\cite{luo2024grounded,luo2022learning,nagarajan2019grounded,li2023locate}, \textit{learning object affordances from human-object interaction images or videos without using any pixel-level affordance annotations.}~As shown in Fig~\ref{fig:1}, given exocentric human-object interaction images and object images with corresponding interaction labels (\eg, lie on) during training, the goal in inference is to ground the affordances of each interaction label on the target egocentric object image.

Two essential yet challenging cores of affordance grounding are \textbf{\textit{(1) extracting affordance knowledge from exocentric interactions}} and \textbf{\textit{(2) transferring it to egocentric localization.}} 
\textbf{\textit{On one hand, for affordance knowledge extraction}}, current methods~\cite{li2023locate,xu2024weakly,jang2024intra} primarily \textit{rely solely on exocentric interaction images,} using CAM~\cite{cam2016} to generate activation maps. CAM is a classic approach for localizing category-relevant regions using image-level labels by highlighting the most discriminative areas for classification. These methods extract and represent interaction knowledge using image features from the generated activation maps, as shown in the ``exocentric+KE" branch in Fig~\ref{fig:2}\textcolor{cvprblue}{a-b}. 
However, the diversity and complexity of interaction scenes make exocentric images alone insufficient for precise affordance activation, leading to vague, broad regions that may include background or human body parts in simple scenarios, and scattered attention in more complex interactions (see Appendix). 
\textbf{\textit{On the other hand, transferring knowledge to egocentric localization}} presents notable challenges due to \textit{the significant domain gap between exocentric and egocentric images, as well as the increased difficulty of transferring knowledge from occluded objects in interaction regions.}  
First, exocentric images are often cluttered with small and potentially occluded interaction regions, while egocentric images are clear and object-centered. Most methods~\cite{luo2022learning, xu2024weakly} that constrain appearance similarity of affordance regions between views fail to address knowledge transfer in complex interaction scenes with large domain gaps, often resulting in mislocalization of non-affordance object parts, as illustrated in the ``egocentric+AL" branch in Fig~\ref{fig:2}\textcolor{cvprblue}{a}. 
Second, while recent advances~\cite{li2023locate} propose part selection (Fig~\ref{fig:2}\textcolor{cvprblue}{b}) on egocentric images to mitigate partial occlusion issues, they still rely on exocentric interaction appearance for guidance, limiting applicability in fully occluded interactions, such as ``lie on" and ``ride," as shown in Fig~\ref{fig:1}.

In this paper, we propose a novel closed-loop knowledge transfer framework, LoopTrans, to address these challenges. Unlike the conventional one-way, non-closed-loop pipeline from exocentric interaction to activation and then to egocentric localization (see the arrow flow in Fig~\ref{fig:2}\textcolor{cvprblue}{a} and \textcolor{cvprblue}{2b}), LoopTrans introduces an innovative closed-loop mechanism that enables egocentric localization to feed back into knowledge attention (Fig~\ref{fig:2}\textcolor{cvprblue}{c}). LoopTrans's dual design naturally addresses key challenges: (1) Egocentric localization on simple, object-centered egocentric images provides clear localization without interference from the background or human body. It can naturally be used to refine knowledge activation in complex exocentric images (see \textcolor{ouryellow}{yellow} arrow flow in Fig~\ref{fig:2}\textcolor{cvprblue}{c}), focusing more precisely on potential affordance regions and resolving coarse and scattered activation issues. (2) In turn, the shared and unified affordance knowledge activation learning across exocentric and egocentric modalities (see \textcolor{ourgreen}{green} arrow flow in Fig~\ref{fig:2}\textcolor{cvprblue}{c}) not only reduces the domain gap but also enables direct use of egocentric activation to transfer affordance knowledge (see \textcolor{ourred}{red} arrow flow in Fig~\ref{fig:2}\textcolor{cvprblue}{c}), bypassing exocentric-to-egocentric appearance transfer and naturally overcoming transfer difficulties caused by occlusions in egocentric interactions. 

Specifically, LoopTrans works as follows in a closed-loop process. 
First, interaction \textcolor{ourgreen}{\textbf{→}} activation: Based on image-level affordance labels, LoopTrans learns a unified classifier for both egocentric and exocentric images, using a shared CAM to highlight both their affordance knowledge activation. Exocentric images provide interaction knowledge, while egocentric images help CAM activation focus more on the object, reducing background and human body interference. More importantly, this shared CAM can effectively identify affordance regions in egocentric objects even when interactions in exocentric images are fully occluded.
Second, activation \textcolor{ourred}{\textbf{→}} localization: Thanks to the shared knowledge attention, LoopTrans refines egocentric localization directly through egocentric activation, rather than relying on exocentric-to-egocentric appearance transfer as in previous methods, thereby reducing the challenges of cross-domain knowledge transfer. LoopTrans directly selects clustered object parts in egocentric images based on egocentric activation, refining coarse activation regions into precise affordance localization. 
Third, localization \textcolor{ouryellow}{\textbf{→}} activation: LoopTrans leverages more precise localization to improve knowledge activation for both egocentric and exocentric images. 
\textcolor{black}{
Given the noisier background of an exocentric image compared to an egocentric one, we introduce a novel denoising distillation method. Using exocentric activation as an anchor, we align the egocentric activation with this anchor while distancing its noise, thereby sharpening the focus on more precise and complete affordance regions and effectively segregating background noise.}
These three components form the end-to-end LoopTrans, establishing a closed-loop knowledge transfer from activation to localization and back, enhancing the precision and transfer of affordance knowledge. 
\vspace{+1mm}

\noindent In summary, our main contributions are multi-fold:
\begin{itemize}
    \item We are the first to propose a closed-loop knowledge transfer mechanism for affordance grounding, where exocentric knowledge activation and egocentric localization mutually enhance each other in a closed loop. 
    \item We propose a shared CAM that enables unified knowledge activation using both exocentric and egocentric images. It not only leverages object-centered egocentric images for clearer activation but also addresses challenges in cross-domain transfer. 
    \item We introduce a novel denoising distillation mechanism that transfers egocentric localization back into the shared CAM, reducing the impact of background noise from exocentric images and focusing activation on object regions. 
    \item Our LoopTrans achieves significant and consistent improvements in weakly-supervised affordance grounding across all metrics on both image and video benchmarks, demonstrating its effectiveness and robustness. 
\end{itemize}
\vspace{-0.3mm}
\section{Related Work}
\label{sec:related_work}
\vspace{-0.7mm}

\begin{figure*}[t]
    \centering
    \includegraphics[width=0.98\linewidth]{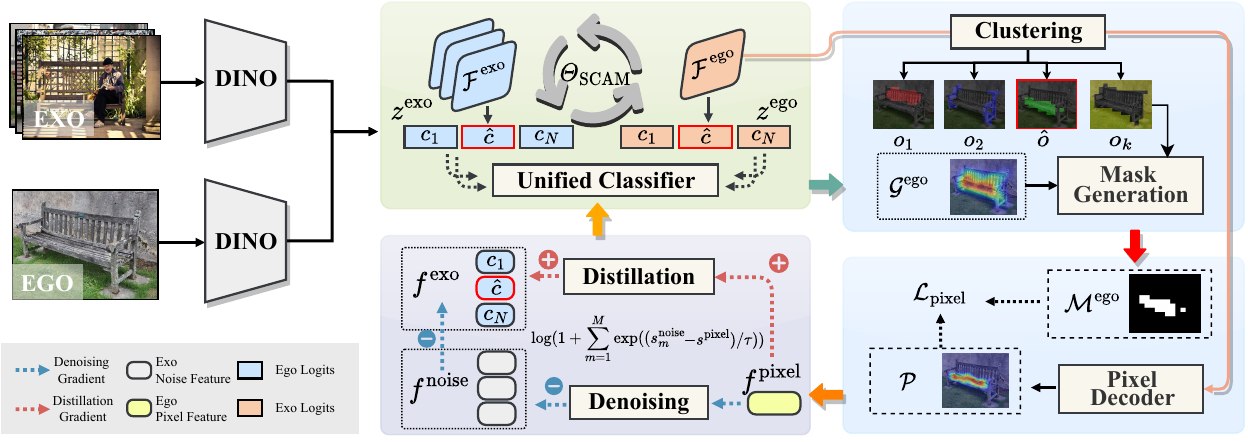}
    \vspace{-2mm}
    \caption{Overall framework of our proposed LoopTrans enables closed affordance knowledge transfer, with the process interaction \textcolor{ourgreen}{\textbf{→}} activation \textcolor{ourred}{\textbf{→}} localization \textcolor{ouryellow}{\textbf{→}} activation, spanning both exocentric and egocentric domains.}
    \vspace{-4mm}
    \label{fig:3}
\end{figure*}

\noindent \textbf{Learning to Localize from Weakly Supervision.}
Weakly supervised localization tasks rely on image-level labels or keypoint annotations to guide object localization and segmentation. Class Activation Map (CAM)~\cite{cam2016}, a foundational approach, highlights discriminative regions linked to image-level labels but often misses less salient areas. To address this, methods have incorporated augmented training~\cite{xu2022cream,mai2020erasing}, semantic~\cite{zhang2018adversarial,kim2022bridging,gao2021ts} and spatial priors~\cite{pan2021unveiling} to improve completeness and accuracy of highlighted regions.
Recently, weak supervision has been extended to affordance localization, where models learn from exocentric images or videos of human-object interactions and transfer activation maps to egocentric images for localization~\cite{luo2022learning,li2023locate}. Despite using interaction data, the task remains challenging due to the diversity of exocentric scenes. Some studies decompose interactions into shared affordance features and individual biases~\cite{luo2022learning, luo2024grounded}, extract hand actions to support affordance inference~\cite{luo2023learning}, or leverage knowledge priors like CLIP~\cite{radford2021clip} to refine affordance grounding with text-based cues~\cite{xu2024weakly}.

\noindent \textbf{Visual Affordance Grounding} focuses on identifying image or video regions likely corresponding to specific human interactions. Early methods relied on small datasets with pixel-level annotations~\cite{chuang2018learning,do2018affordancenet,koppula2013learning,myers2015affordance}, using object geometry~\cite{myers2015affordance} or appearance~\cite{koppula2013learning} to infer affordances. More recent approaches favor weakly supervised methods, which are more feasible in real-world scenarios. Studies~\cite{sawatzky2017adaptive} demonstrate that effective affordance localization can be achieved using minimal keypoint annotations. Later works~\cite{nagarajan2019grounded,luo2022learning} incorporate human-object interaction (HOI) priors, relying solely on affordance category labels to reduce annotation costs. However, the diversity and complexity of HOI scenes introduce new challenges for model learning. 
To address these, ~\cite{luo2023learning} uses hand cues to reduce action ambiguity, while ~\cite{luo2022learning} decomposes interactions to capture shared affordances across diverse contexts. Approaches like ~\cite{xu2024weakly} and ~\cite{li2023locate} employ localized knowledge transfer to filter out irrelevant backgrounds, with ~\cite{li2023locate} further adding a part selection to isolate specific object-part features. Recognizing the limitations of traditional activation maps in complex scenes, ~\cite{xu2024weakly} and~\cite{jang2024intra,qian2024affordancellm} incorporate auxiliary activations from CLIP and large language models for improved support.
\section{Preliminary}
\label{sec:3.1}
\noindent \textbf{Problem Definition.} 
Given pairs of exocentric and egocentric images with image-level affordance labels, weakly supervised affordance grounding~\cite{luo2024grounded,luo2022learning,nagarajan2019grounded} aims to extract interaction knowledge from exocentric images and accurately locate object parts corresponding to affordance labels in egocentric images. 
Specifically, given a pair of exocentric and egocentric images, \(\{I^{\text{exo}}, I^{\text{ego}}\}\) along with \(N\) affordance categories \(\{c_1, \ldots, c_N\}\), the objective is to generate egocentric affordance activation maps \(\mathcal{G}^{\text{ego}}\in \mathbb{R}^{H\times W\times N}\), corresponding to the $N$ categories. Here, \(H\) and \(W\) are the height and width of feature maps, respectively. 

\noindent\textbf{Visual Feature Extraction.} Self-supervised ViT DINO~\cite{dino2021} is employed to extract image patch features, denoted as \( \mathcal{F}^{\text{exo}} \in \mathbb{R}^{H\times W \times C}\) and \( \mathcal{F}^{\text{ego}} \in \mathbb{R}^{H\times W \times C}\), where $C$ is the feature dimension. The interaction-focused exocentric feature \(\mathcal{F}^{\text{exo}}\) encodes affordance knowledge by highlighting interaction regions and types, while the object-centered egocentric feature \(\mathcal{F}^{\text{ego}}\) concentrates on object-specific information without background and human interference.

\noindent\textbf{Class Activation Mapping (CAM)}~\cite{cam2016} provides a mechanism for localizing affordance activation regions through image-level weakly-supervised learning. Given an input feature map \( \mathcal{F}\in \mathbb{R}^{H \times W \times C} \), it is first transformed via an MLP followed by a two-layer convolutional block. A subsequent \(1\times1\) convolutional layer with \(N\) category-specific kernels generates activation maps \(\mathcal{G}\in \mathbb{R}^{H\times W\times N}\), where each kernel's output corresponds to the activation region for a specific affordance category. Activation maps undergo global pooling to generate activation scores for final class probability prediction. The CAM process is defined as:
\vspace{-1mm}
\begin{equation}
\begin{aligned}
&\mathcal{G} = \varTheta_{\text{CAM}}(\mathcal{F};\theta), z = \operatorname{GAP}(\mathcal{G}), \\
&\mathcal{L}_{\text{cls}} = - \scalebox{1.1}{$\sum$}_{i=1}^N \mathbb{I}(c_i=\hat{c}) \log \sigma(z_i),
\end{aligned}
\end{equation}
where \(\varTheta_{\text{CAM}}\) represents the CAM module, \(\theta\) is the trainable parameters of CAM, \(\operatorname{GAP}(\cdot)\) denotes global average pooling,  \(z \in \mathbb{R}^{N}\) represents activation scores, with class probabilities from the sigmoid function \(\sigma\). \(\mathbb{I}\) is the indicator function, and \(\mathcal{L}_{\text{cls}}\) is the classification loss designed to align the predicted probability with the ground-truth label $\hat{c}$. By directly minimizing this classification loss based on the activation scores, its corresponding activation maps effectively highlight the regions of the affordance category.

\noindent \textbf{One-Way Exo-Ego Transfer.} As affordance-based knowledge sources are exocentric and localization targets are egocentric, spanning different image domains, existing methods have predominantly employed a one-way framework that aligns exocentric features to egocentric ones for grounding~\cite{luo2023learning, luo2024grounded, luo2022learning, nagarajan2019grounded, li2023locate, xu2024weakly, jang2024intra}.
They use two independent CAM modules to activate affordance regions in exocentric and egocentric images separately, then align the features corresponding to the two activation regions. 
Specifically, for paired images \(I^{\text{exo}}\) and \(I^{\text{ego}}\) sharing the same category \(\hat{c}\), paired CAM modules with distinct parameters \(\theta^{\text{exo}}\) and \(\theta^{\text{ego}}\) are employed to produce the corresponding activation maps \(\mathcal{G}^{\text{exo}}\) and \(\mathcal{G}^{\text{ego}}\). The activation maps are then combined with feature maps through weighted averaging to obtain corresponding features $h^{\text{exo}}$ and $h^{\text{ego}}$, which are finally aligned to achieve one-way transfer. The entire computational process is formulated as follows:
\vspace{-1mm}
\begin{equation}
\begin{aligned}
    &h^{\text{exo}} = \operatorname{GAP}\left(\mathcal{R}(\mathcal{G}^{\text{exo}}_{\hat{c}}) \circ \mathcal{F}^{\text{exo}}\right), \\ 
    &h^{\text{ego}} = \operatorname{GAP}\left(\mathcal{R}(\mathcal{G}^{\text{ego}}_{\hat{c}}) \circ \mathcal{F}^{\text{ego}}\right), \mathcal{L}_{\text{align}} = \|h^{\text{exo}} - h^{\text{ego}}\|^2_2, 
\end{aligned}
\end{equation}
where \(\mathcal{R}(\cdot)\) and \(\circ\) denote min-max normalization and the Hadamard product, respectively. \(\mathcal{G}^{\text{exo}}_{\hat{c}}\) denotes the activation map of class \(\hat{c}\) in \(\mathcal{G}^{\text{exo}}\), and $\|\cdot\|_2$ means L2 norm. 

\vspace{-0.3mm}
\section{Method}
\label{sec:method}
\vspace{-0.7mm}
\noindent \textbf{Overview.} Existing weakly supervised affordance grounding methods face two critical limitations:
(1) one-way feature alignment heavily depends on exocentric features. However, exocentric activation regions often include human hands or body parts, blending object information with background elements. This prevents egocentric activation features and maps from focusing solely on the object, while isolated CAM modules further amplify view discrepancies. 
(2) one-way exocentric-to-egocentric transfer underutilizes the object-centric nature of egocentric images, which could also aid exocentric activation. A bidirectional transfer enhances activation consistency and mitigates the domain gap between context-rich exocentric and object-centric egocentric images. 
Therefore, we propose \textbf{LoopTrans}, a closed-loop framework (Fig.~\ref{fig:3}), which facilitates the knowledge transfer in a loop, including three key stages:
\begin{itemize}
\item \textbf{Interaction \textcolor{ourgreen}{\textbf{→}} Activation.} We achieve shared interaction knowledge activation and transfer through a unified CAM module jointly trained on both views (Sec~\ref{sec:3.3}). Our shared activation allows exocentric interaction patterns to directly activate object-centric affordance regions in egocentric images, effectively eliminating background interference caused by explicit feature alignment.
\item \textbf{Activation \textcolor{ourred}{\textbf{→}} Localization.} 
Leveraging egocentric images' object-centric nature, we use DINO feature clustering to extract object parts and train a pixel decoder with part-level pseudo-masks to refine coarse activation maps from the previous stage into precise, egocentric affordance localization (Sec~\ref{sec:3.4}). 
\item \textbf{Localization \textcolor{ouryellow}{\textbf{→}} Activation.}
Refined egocentric localization from the second stage is fed back to enhance shared knowledge activation in the first stage, eliminating irrelevant context and directing activation toward affordance-relevant objects via our denoising distillation (Sec~\ref{sec:3.5}).
\end{itemize}

\vspace{-0.3mm}
\subsection{Unified Exo-to-Ego Activation}
\label{sec:3.3}
\vspace{-0.7mm}

In this section, we aim to extract interaction knowledge to jointly highlight activation maps from exocentric and egocentric images. Instead of previous one-way grounding methods that process exocentric \(\{I^{\text{exo}}\}\) and egocentric \(\{I^{\text{ego}}\}\) images separately using distinct CAM modules, we propose \(\varTheta_{\text{SCAM}}\), a Shared CAM module that unifies exo-ego activation through parameter sharing and co-training.

\begin{figure}[t]
    \centering
    \includegraphics[width=0.98\linewidth]{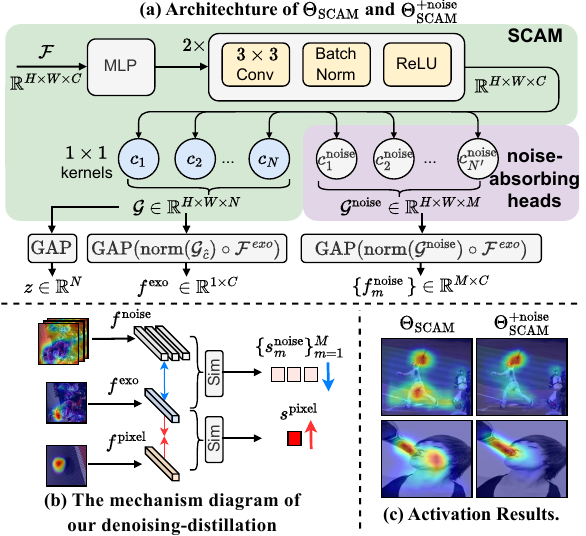}
    \vspace{-2mm}
    \caption{(a) Network architectures of \(\varTheta_{\text{SCAM}}\) and our proposed \(\varTheta_{\text{SCAM}}^{\text{+noise}}\); (b) \textcolor{black}{The mechanism} of our denoising distillation; (c) \textcolor{black}{Exocentric activation results of} \(\varTheta_{\text{SCAM}}\) and \(\varTheta_{\text{SCAM}}^{\text{+noise}}\).}
    \vspace{-6mm}
    \label{fig:6}
\end{figure}

Specifically, while preserving the vanilla CAM architecture~\cite{cam2016} (as illustrated in Fig~\ref{fig:6}\textcolor{cvprblue}{a} \colorbox{scamgreen}{green}), \(\varTheta_{\text{SCAM}}\) processes both exocentric features \(\mathcal{F}^{\text{exo}}\) and egocentric features \(\mathcal{F}^{\text{ego}}\) using identical learnable parameters $\theta$, as follows:
\vspace{-2mm}
\begin{equation}
\begin{aligned}
&\mathcal{G}^{\text{exo}}, \mathcal{G}^{\text{ego}} = \varTheta_{\text{SCAM}}\left( \{\mathcal{F}^{\text{exo}}, \mathcal{F}^{\text{ego}}\}; \theta \right), \\
&z^{\text{exo}}, z^{\text{ego}} = \operatorname{GAP}(\mathcal{G}^{\text{exo}}), \operatorname{GAP}(\mathcal{G}^{\text{ego}}), \\
&\mathcal{L}_{\text{cls}} = -\scalebox{1.1}{$\sum$}_{i=1}^{N} \mathbb{I}(c_i=\hat{c}) \log\left(\sigma(z^{\text{exo}}_i) \cdot \sigma(z^{\text{ego}}_i)\right),
\end{aligned}
\vspace{-2mm}
\end{equation}
where shared parameters \(\theta\) enforce cross-view consistency. The classification loss \(\mathcal{L}_{\text{cls}}\) maximizes joint confidence \(\sigma(z^{\text{exo}}_i) \cdot \sigma(z^{\text{ego}}_i)\) for class \(\hat{c}\), driving \(\varTheta_{\text{SCAM}}\) to align affordance predictions across views. This bidirectional synergy allows exocentric activations to suppress human-body interference using egocentric object cues, while egocentric maps implicitly acquire affordance interaction knowledge from exocentric interaction context.

Notably, to further mitigate background and human noise in exocentric images during shared activation, we introduce multiple noise-absorbing heads into the final activation convolutional layer, as illustrated in Fig \ref{fig:6}\textcolor{cvprblue}{a} \colorbox{scampurple}{purple}. These heads isolate non-affordance context and background noise for exocentric images, effectively reducing domain gap-induced interference in cross-view shared activations. The detailed implementation and functionality of these heads are elaborated in Section~\ref{sec:3.5}. 

\subsection{Region Activation to Pixel Localization}
\label{sec:3.4}
Leveraging our unified knowledge extraction through shared CAM, LoopTrans can directly generate the egocentric activation map \(\mathcal{G}^{\text{ego}}\), thereby alleviating the challenges associated with cross-domain knowledge transfer. However, both weakly-supervised object localization and affordance grounding tasks have long faced a key challenge~\cite{zhang2018adversarial,gao2021ts,lee2021railroad,luo2022learning}: CAM highlights only the most salient regions, resulting in activation maps that inadequately cover the entire interaction part. To address this, we transform the coarse and even ambiguous knowledge activation maps into clearly defined object part pseudo-masks (Sec~\ref{sec:3.4.1}). Furthermore, we train a pixel-level affordance decoder \(\varTheta_{\text{pixel}}\) that utilizes the pseudo-masks to generate accurate affordance localization. (Sec~\ref{sec:3.4.2}).

\subsubsection{Activation to \textit{Object Part}}
\label{sec:3.4.1}
We leverage the properties of the self-supervised ViT DINO~\cite{dino2021} to partition egocentric images into semantically distinct parts through unsupervised clustering, subsequently generating pseudo-labels that encompass complete object parts based on the activation map \(\mathcal{G}^{\text{ego}}\). Specifically, given the egocentric image feature \(\mathcal{F}^{\text{ego}}\) that mainly contains objects, we first apply unsupervised clustering to divide them into \(K\) parts \(\{o_1, \ldots, o_K\}\), where \(o_k \in \{0,1\}^{H \times W}\). Each part \(o_k\) is assigned a clear and distinct semantic region, such as bench \textbf{→} backrest, armrest, seat, and background, as illustrated in the top right corner of Fig~\ref{fig:3}. Next, we select the part with the highest Intersection over Union (IoU) score relative to the activation map \(\mathcal{G}^{\text{exo}}\) as the accurate localization result, which serves as the pseudo mask \(\mathcal{M}^{\text{exo}}\). The detailed computational process is as follows:
\vspace{-1mm}
\begin{equation}
\begin{aligned}
&\mathcal{M}^{\text{ego}} = \hspace{-5pt}\mathop{\arg\max}\limits_{o_k\in\{o_1, \cdots, o_K\}}\hspace{-5pt} \operatorname{IoU}\left(o_k, \mathbb{I}(\mathcal{R}(\mathcal{G}_{\hat{c}}^{\text{ego}}) \geq \mu)\right),
\end{aligned}
\vspace{-1mm}
\end{equation}
where \( \mathcal{G}_{\hat{c}}^{\text{ego}}\) means the activation map for class $\hat{c}$, and \( \mu \) is the threshold used to filter foreground, and \( \mathbb{I}(\cdot \geq \mu) \) is an indicator function that returns 1 when the value is greater than or equal to the threshold \( \mu \), and 0 otherwise.

\subsubsection{\textit{Object Part} to Localization}
\label{sec:3.4.2}
The objective of this section is to train a pixel-level affordance decoder, \(\varTheta_{\text{pixel}}\), which employs region-complete pseudo mask, \(\mathcal{M}^{\text{ego}}\), as supervision to learn the final affordance localization in exocentric images. 
\textcolor{black}{$\varTheta_{\text{pixel}}$ has the same architecture with $\varTheta_{\text{CAM}}$ but has its own learnable parameters.} 
Given the feature \(\mathcal{F}^{\text{ego}}\) of the input egocentric image and the corresponding pseudo mask \(\mathcal{M}^{\text{ego}}\), we feed \(\mathcal{F}^{\text{ego}}\) into \(\varTheta_{\text{pixel}}\) to obtain the per-pixel localization map \(\mathcal{P} \in \mathbb{R}^{H \times W \times N}\). 
Here, \(N\) represents the number of affordance categories, \(0 \leq \mathcal{P}_{i,j,c} \leq 1\) represents the probability that the pixel at position \((i,j)\) of the image corresponds to the localization of the \(c\)-th affordance class, and we denote \(\mathcal{P}_{\hat{c}} \in \mathbb{R}^{H \times W}\) as the probability map corresponding to affordance category \(\hat{c}\). We supervise the pixel-level affordance decoder using \(\mathcal{L}_{\text{pixel}}\), which combines dice~\cite{li2019dice} with MSE losses, as follows:
\begin{equation}
\begin{aligned}
\mathcal{P} = &\varTheta_{\text{pixel}}(\mathcal{F}^{\text{ego}};\theta), \mathcal{L}_{\text{mse}}=\scalebox{1.1}{$\frac{1}{HW}$}||\mathcal{P}_{\hat{c}}-\mathcal{M}^{\text{ego}}||^2,\\ 
&\mathcal{L}_{\text{dice}} = 1-\frac{2\sum_{i,j}\mathcal{P}_{i,j,\hat{c}} \cdot \mathcal{M}^{\text{ego}}_{i,j}}{\sum_{i,j}\mathcal{P}_{i,j,\hat{c}}+\sum_{i,j}\mathcal{M}^{\text{ego}}_{i,j}}.
\end{aligned}
\vspace{-1mm}
\end{equation}

With per-pixel supervision, LoopTrans achieves precise and complete localization region for affordances. More importantly, comprehensive supervision at the object-region level ensures consistency in knowledge transfer across images, fundamentally addressing the information asymmetry during transferring caused by domain difference between exocentric and egocentric.

\subsection{Ego-to-Exo Denoising Distillation}
\label{sec:3.5}  
Precise affordance activation in exocentric images is hindered by scene complexity and small object scales, where conventional CAMs tend to over-activate human-centric regions while overlooking subtle interaction cues (Fig~\ref{fig:6}\textcolor{cvprblue}{c}). To address this limitation, we introduce the denoising distillation that reverse-propagates pixel-level affordance priors from egocentric to exocentric views, thereby refining interaction knowledge by effectively suppressing background and human interference.

As shown in Fig~\ref{fig:6}\textcolor{cvprblue}{a-b}, our denoising distillation mechanism enhances the SCAM module by integrating parallel noise-absorbing heads, forming $\varTheta_{\text{SCAM}}^{\text{+noise}}$. This extension aims to explicitly isolate non-affordance patterns (\eg, human limbs, cluttered backgrounds) from exocentric features. Given an exocentric feature map \(\mathcal{F}^{\text{exo}}\), we simultaneously generate the primary affordance activation \(\mathcal{G}^{\text{exo}}\) and \(M\) noise-specific activations \(\mathcal{G}^{\text{noise}}\in \mathbb{R}^{H\times W\times M}\) through additional \(M\) convolution kernels:
\vspace{-1mm}
\begin{equation}
\begin{aligned}
&\mathcal{G}^{\text{exo}}, \mathcal{G}^{\text{noise}} = \varTheta_{\text{SCAM}}^{\text{+noise}}(\mathcal{F}^{\text{exo}};\theta^{\text{+noise}}) \\
&f^{\text{exo}} = \operatorname{GAP}(\mathcal{R}(\mathcal{G}^{\text{exo}}_{\hat{c}}) \circ \mathcal{F}^{\text{exo}})\\
&f^{\text{pixel}}=\operatorname{GAP}(\mathcal{R}(\mathcal{P}_{\hat{c}}) \circ \mathcal{F}^{\text{ego}}), \\
&\{f^{\text{noise}}_m\} =  \operatorname{GAP}\left(\mathcal{R}(\{\mathcal{G}^{\text{noise}}_m\}) \circ \mathcal{F}^{\text{exo}}\right),
\end{aligned}
\vspace{-1mm}
\end{equation}  
where \textcolor{black}{\(\theta^{\text{+noise}}\) denotes the parameters with additional noise-absorbing heads, \(\mathcal{P}_{\hat{c}}\) represents the egocentric localization result  corresponding to the target class \(\hat{c}\). The Hadamard product \(\circ\), followed by global average pooling (\(\operatorname{GAP}\)), is used to extract exocentric affordance activation features \( f^{\text{exo}} \), noise features \( f_m^{\text{noise}}  \), and egocentric localization feature \( f^{\text{pixel}} \).}
\begin{table*}[!htb]
    \centering
    \small
    \setlength\tabcolsep{2.5mm}
    \renewcommand\arraystretch{0.9}
    \begin{tabular}{llccccccccc}
        \toprule
        \multirow{2.5}{*}{\textbf{Method}} & \multirow{2.5}{*}{\textbf{Pub.}} & \multicolumn{3}{c@{\hskip 0.1in}}{\textbf{AGD20K-Seen}} & \multicolumn{3}{c@{\hskip 0.1in}}{\textbf{AGD20K-Unseen}} & \multicolumn{3}{c@{\hskip 0.1in}}{\textbf{HICO-IFF}} \\
        \cmidrule(r){3-5} \cmidrule(r){6-8} \cmidrule(r){9-11}
            &   & KLD$\downarrow$ & SIM$\uparrow$ & NSS$\uparrow$ & KLD$\downarrow$ & SIM$\uparrow$ & NSS$\uparrow$ & KLD$\downarrow$ & SIM$\uparrow$ & NSS$\uparrow$ \\
        \midrule
        \multicolumn{10}{l}{\textbf{Weakly Supervised Object Localization}} \\
        SPA~\cite{pan2021spa} & CVPR21 & 5.528 & 0.221 & 0.357 & 7.425 & 0.169 & 0.262 & --- & --- & --- \\
        EIL~\cite{mai2020eil} & CVPR20 & 1.931 & 0.285 & 0.522 & 2.167 & 0.277 & 0.330 & --- & --- & --- \\
        TS-CAM~\cite{gao2021ts} & ICCV21 & 1.842 & 0.260 & 0.336 & 2.104 & 0.201 & 0.151 & --- & --- & --- \\
        \midrule
        \multicolumn{10}{l}{\textbf{Affordance Grounding}} \\
        Hotspots~\cite{nagarajan2019grounded} & ICCV19 & 1.773 & 0.278 & 0.615 & 1.994 & 0.237 & 0.577 & --- & --- & --- \\
        Cross-view-AG~\cite{luo2022learning} & CVPR22 & 1.538 & 0.334 & 0.927 & 1.787 & 0.285 & 0.829 & 1.779 & 0.263 & 0.946 \\
        Cross-view-AG+~\cite{luo2024grounded} & --- & 1.489 & 0.342 & 0.981 & 1.765 & 0.279 & 0.882 & 1.836 & 0.256 & 0.883 \\
        LOCATE~\cite{li2023locate} & CVPR23 & 1.226 & 0.401 & 1.177 & 1.405 & 0.372 & 1.157 & 1.593 & 0.327 & 0.966\\
        WSMA~\cite{xu2024weakly} & AAAI24 & \underline{1.176} & \underline{0.416} & \underline{1.247} & \underline{1.335} & \underline{0.382} & \underline{1.220} & \underline{1.465} & \underline{0.358} & \underline{1.012}\\
        INTRA~\cite{jang2024intra} & ECCV24 & 1.199 & 0.407 & 1.239 & 1.365 & 0.375 & 1.209 & --- & --- & ---\\
        \textbf{Ours} & --- & \textbf{1.088} & \textbf{0.445} & \textbf{1.322} & \textbf{1.247} & \textbf{0.403} & \textbf{1.315} & \textbf{1.399} & \textbf{0.379} & \textbf{1.226} \\
        \bottomrule
    \end{tabular}
    \vspace{-2mm}
    \caption{Comparison results on AGD20K-Seen, AGD20K-Unseen, and HICO-IFF benchmarks, where interaction knowledge is derived from exocentric images. The highest performance is \textbf{bolded}, and the second-highest is \underline{underline}.}
    \label{tab:1}
    \vspace{-2mm}
\end{table*}

To ensure \( f_m^{\text{noise}} \) captures noise effectively while directing affordance activation \(\mathcal{G}^{\text{exo}}_{\hat{c}}\) toward affordance regions, we introduce a denoising distillation mechanism. 
The core idea is to align exocentric affordance-related features \( f^{\text{exo}} \) with clean egocentric object features \( f^{\text{pixel}} \), while enforcing noise features \( f^{\text{noise}} \) to diverge from affordance-related features \( f^{\text{exo}} \). This naturally pushes noise activation toward object-irrelevant contexts and background regions. The denoising distillation is as follows:
\vspace{-1mm}
\begin{equation}
\begin{aligned}
    &s_m^{\text{noise}},s^{\text{pixel}} = \operatorname{sim}(f_m^{\text{noise}}, f^{\text{exo}}), \operatorname{sim}(f^{\text{pixel}}, f^{\text{exo}})\\
    &\mathcal{L}_{\text{dill}} = \log(1+\scalebox{1.1}{$\sum$}_{m=1}^M\operatorname{exp}((s^{\text{noise}}_m-s^{\text{pixel}})/\tau)),
\end{aligned}
\vspace{-1mm}
\end{equation}
\textcolor{black}{where \(\operatorname{sim}(a, b)\) denotes the cosine similarity calculation, \(s_m^{\text{noise}} \in \mathbb{R}^{1}\) represents the similarity between the noise features of the \(m\)-th noise-absorbing head and the exocentric activation features, and \(s^{\text{pixel}} \in \mathbb{R}^{1}\) denotes the similarity between the egocentric localization features and the exocentric activation features. The loss function encourages the exocentric activation features to align with the precise localization features from the pixel decoder while penalizing the similarity between the affordance of exocentric images and irrelevant background features.}

Finally, the overall loss for LoopTrans is defined as: 
\begin{equation}
\mathcal{L} = \lambda_{\text{cls}}\mathcal{L}_{\text{cls}} + \lambda_{\text{dill}}\mathcal{L}_{\text{dill}} + \lambda_{\text{pixel}}\mathcal{L}_{\text{pixel}} + \lambda_{\text{corr}}\mathcal{L}_{\text{corr}},
\end{equation}
where $\lambda_{\text{cls}}, \lambda_{\text{dill}}, \lambda_{\text{pixel}}, \lambda_{\text{corr}}$ represents the weights assigned to the different loss components. $\mathcal{L}_{\text{corr}}$ is used to \textcolor{black}{align the affordance correlations between exocentric and egocentric images} following \cite{xu2024weakly,luo2022learning}. The entire model is trained in an end-to-end manner.
\begin{table*}[h]
\centering
\small
\setlength\tabcolsep{3.5mm}
\renewcommand\arraystretch{0.8}
\begin{tabular}{clcccccc}
\toprule
&\multirow{2.5}{*}{\textbf{Method}} & \multicolumn{3}{c@{\hskip 0.1in}}{\textbf{EPIC}} & \multicolumn{3}{c@{\hskip 0.1in}}{\textbf{OPRA}}  \\
 \cmidrule(r){3-5} \cmidrule(r){6-8}
 &  & KLD $\downarrow$ & SIM $\uparrow$ & AUC-J $\uparrow$& KLD $\downarrow$ & SIM $\uparrow$ & AUC-J $\uparrow$\\
\midrule
\multirow{3}{*}{\textbf{Supervised}}  
& Img2heatmap~\cite{nagarajan2019grounded} & 1.400 & 0.359 & 0.794 & 1.473 & 0.355 & 0.821 \\
& Demo2Vec~\cite{fang2018demo2vec} & --- & --- & --- & 1.197 & 0.482 & 0.847 \\
& Afformer~\cite{chen2023affordance} & 0.97 & 0.56 & 0.88 & 1.05 & 0.53 & 0.89 \\
\midrule
\multirow{9}{*}{\textbf{\shortstack{Weakly\\Supervised}}} 
& Hotspot$^\dag$~\cite{nagarajan2019grounded} &--- &--- &--- & \textcolor{gray}{1.427} & \textcolor{gray}{0.362} & \textcolor{gray}{0.806} \\
& HAG-Net$^\dag$~\cite{luo2023learning} &--- &--- &--- & \textcolor{gray}{1.409} & \textcolor{gray}{0.365} & \textcolor{gray}{0.812} \\
\cmidrule(r){2-8}
& MLNET~\cite{cornia2016mlnet} & 6.116 & 0.318 & 0.746 & 4.022 & 0.284 & 0.763 \\
& EGOGAZE~\cite{huang2018egogaze} & 2.241 & 0.273 & 0.614 & 2.428 & 0.245 & 0.646 \\
& SALGAN~\cite{pan2017salgan} & 1.508 & 0.395 & 0.774 & 2.116 & 0.309 & 0.769 \\
& DEEPGAZEII~\cite{kummerer2016deepgaze} & 1.352 & 0.394 & 0.751 & 1.897 & 0.296 & 0.720 \\
& Hotspot~\cite{nagarajan2019grounded} & 1.258 & 0.404 & 0.785 & \underline{1.537} & \underline{0.342} & \underline{0.754} \\
& HAG-Net~\cite{luo2023learning} & \underline{1.209} & \underline{0.414} & \underline{0.801} &--- &--- &--- \\
& \textbf{Ours} & \textbf{1.130} & \textbf{0.431} & \textbf{0.827} & \textbf{1.429} & \textbf{0.358} & \textbf{0.804} \\
\midrule
\multirow{3}{*}{\textbf{\shortstack{Image-to-Video\\genralization}}} 
& LOCATE~\cite{li2023locate} & \underline{1.382} & \underline{0.394} & 0.668 & 1.620 & 0.342 & 0.682 \\
& WSMA~\cite{xu2024weakly} & 1.425 & 0.371 & \underline{0.720} & \underline{1.536} & \underline{0.344} & \underline{0.748} \\
& \textbf{Ours} & \textbf{1.244} & \textbf{0.405} & \textbf{0.785} & \textbf{1.457} & \textbf{0.355} & \textbf{0.789} \\
\bottomrule
\end{tabular}
\vspace{-2mm}
\caption{Comparison results on EPIC and OPRA benchmarks, where interaction knowledge is derived from exocentric videos.}
\vspace{-5mm}
\label{tab:2}
\end{table*}
\section{Experiments}
\subsection{Datasets and Implementation Details}
\noindent\textbf{Image Benchmarks and Metrics.}
Following previous works~\cite{luo2022learning, li2023locate, xu2024weakly, jang2024intra}, we conduct experiments on AGD20K~\cite{luo2022learning}, which is a large-scale dataset comprising both exocentric and egocentric images and includes the splits ``Seen” and ``Unseen", as well as HICO-IFF derived from HICO-DET~\cite{chao2018learning} and IIT-AFF~\cite{nguyen2017object}. For a fair comparison, we adopt the same metrics as in previous works for performance evaluation: Kullback-Leibler Divergence (KLD), Similarity (SIM), and Normalized Scanpath Saliency (NSS).

\noindent\textbf{Video Benchmarks and Metrics.}
Following works~\cite{nagarajan2019grounded, luo2023learning} in video affordance localization, we evaluate using the OPRA~\cite{fang2018demo2vec} and EPIC-Kitchens~\cite{damen2018scaling} datasets.
Building on our image framework, we integrate an LSTM (after DINO) to represent the exo video using features from its last frame.
Notably, videos of the OPRA dataset are collected from YouTube. However, since some resources are no longer available, results in Table~\ref{tab:2} marked with the $^\dag$ represent experiments conducted with the full dataset, yielding outcomes that are currently unattainable with the accessible subset.
For evaluation, we use the most common metrics in video affordance localization—KLD, SIM, and the Area Under the Curve for the Jaccard index (AUC-J).
For more details on the datasets, please refer to the Appendix. 

\noindent\textbf{Implementation Details.} Following ~\cite{luo2022learning, luo2024grounded, li2023locate, xu2024weakly}, we set the input image resolution to \(224 \times 224\). For video inputs, we sample 8 frames at equal intervals and similarly resize them to \(224 \times 224\). The cluster number $k$ is set to 4. All experiments are conducted on a single NVIDIA TITAN, using SGD as the optimizer with a learning rate of \(1 \times 10^{-3}\).

\subsection{Comparison with State-of-the-Art Methods}

\noindent \textbf{Comparison on Image Benchmarks.}
As shown in Table~\ref{tab:1}, we compare LoopTrans with state-of-the-art methods across three benchmarks (including both splits of AGD20K). Our approach consistently outperforms all existing methods across all evaluation metrics and settings. On AGD20K, we achieve average improvements of \todo{6.7}\% over the the previsou best-performing model WSMA~\cite{xu2024weakly} in KLD, SIM, and NSS, representing a relative increase of \todo{236}\% compared to the gap between prior state-of-the-art methods. Unlike WSMA, which uses additional text domain prompts as a medium for knowledge transfer, our LoopTrans fundamentally addresses localization challenges and inaccuracies in knowledge extraction caused by the domain gap through a unified shared CAM and a reverse egocentric-to-exocentric denoising distillation. Additionally, compared to LOCATE~\cite{li2023locate}, which refines exocentric-to-egocentric transfer through feature selection, our shared CAM demonstrates more accurate knowledge transfer under occlusion and in complex scenes, yielding improvements of \todo{11.3}\% across all three metrics. For HICO-IFF, our consistent improvement of \todo{10.5}\% over WSMA further underscores the effectiveness and adaptability of LoopTrans. 

\noindent \textbf{Comparison on Video Benchmarks.}
As shown in Table~\ref{tab:2}, we comprehensively evaluate the proposed LoopTrans on video datasets from two perspectives: weakly supervised learning and image-to-video generalization. In both settings, LoopTrans consistently outperforms other methods, demonstrating substantial improvements in cross-domain knowledge transfer between exocentric and egocentric domains and significantly enhancing affordance localization accuracy. (1) Weakly Supervised Setting: We compare LoopTrans with approaches that leverage temporal interaction knowledge through dedicated temporal modules. On the EPIC dataset, LoopTrans achieves an average improvement of \todo{4.6}\% over HAG-Net~\cite{luo2023learning}; on the OPRA dataset, it improves by \todo{6.1}\% over Hotspot~\cite{nagarajan2019grounded}. These gains highlight how our unified shared CAM not only bridges the domain gap between exocentric and egocentric perspectives but also supports robust knowledge transfer across video and image modalities, even with substantial modality gaps. (2) Image-to-Video Generalization: Trained on the AGD20K image dataset, LoopTrans is evaluated exclusively on video datasets. Against the current state-of-the-art open-source method, WSMA, LoopTrans achieves a further improvement of about \todo{7.5}\%, underscoring its adaptability and robustness in bridging domain gaps. Notably, our framework performs consistently across exocentric benchmarks without requiring specialized temporal modules.

\begin{table}
    \centering
    \small
    \setlength\tabcolsep{0.77mm}
    \renewcommand\arraystretch{0.72}
    \vspace{2mm}
    \begin{tabular}{c|ccc|ccc}
    \toprule
    & \makecell{\textbf{Unified} \\ \textbf{CAM}} &  \makecell{\textbf{Pixel} \\ \textbf{Alignment}} & \makecell{\textbf{Denoising}\\\textbf{Distillation}} & KLD $\downarrow$ & SIM $\uparrow$ & NSS $\uparrow$  \\
    \midrule
    \multirow{6}{*}{\rotatebox{90}{\textbf{Seen}}} 
    & & & & 1.318  & 0.384 & 1.135 \\
    & $\checkmark$ & & & 1.259 & 0.409 & 1.179\\
    & &  & $\checkmark$ & 1.251 & 0.392 & 1.196 \\
    &  $\checkmark$ & $\checkmark$ &  & 1.149 & 0.425 & 1.266 \\
    &  $\checkmark$ &  & $\checkmark$ & 1.222 & 0.405 & 1.183 \\
    &  $\checkmark$ & $\checkmark$ & $\checkmark$  & \textbf{1.088} & \textbf{0.443} & \textbf{1.322} \\
    \midrule
    
    \multirow{6}{*}{\rotatebox{90}{\textbf{Unseen}}} 
    & & & & 1.635 & 0.332 & 0.853 \\
    & $\checkmark$ & & & 1.508 & 0.341 & 1.122 \\
    & &  & $\checkmark$ & 1.468 & 0.344 & 1.157 \\
    &  $\checkmark$ & $\checkmark$ &  & 1.335 & 0.394 & 1.258 \\
    &  $\checkmark$ &  & $\checkmark$ & 1.431 & 0.368 & 1.189 \\
    &  $\checkmark$ & $\checkmark$ & $\checkmark$ & \textbf{1.247} & \textbf{0.403} & \textbf{1.315} \\
    \bottomrule
    \end{tabular}
    \vspace{-2mm}
    \caption{Ablation study on AGD20K benchmark.}
    \label{tab:3}
    \vspace{-5mm}
\end{table}

\subsection{Ablation Study and Discussions}
Beginning with our baseline, we develop six model variants across two subsets of AGD20K, conducting a total of 12 ablation experiments to rigorously evaluate the contributions of our modules as shown in Table~\ref{tab:3}. 
\textbf{(1) Baseline Model}: We use the one-way affordance grounding pipeline, LOCATE~\cite{li2023locate}, as our baseline without applying feature selection, employing $\mathcal{L}_{\text{corr}}$~\cite{luo2022learning, xu2024weakly} for knowledge transfer.
\textbf{(2) Shared CAM}: Adding shared CAM yields consistent performance improvements (\todo{+4.5\%}) on seen split over KLD. This improvement demonstrates that our shared CAM effectively facilitates knowledge extraction and activation between exocentric and egocentric images, enabling each domain to leverage the other's strengths.
\textbf{(3) Denoising Distillation}: This mechanism enhances baseline performance by \todo{5.1\%} by reinforcing egocentric information back to exocentric data, establishing a closed-loop knowledge cycle that enables CAM to filter out background noise and human-related artifacts while directing attention to interactive objects.
\textbf{(4) Pixel Alignment}: Since pixel alignment requires precise activation maps for egocentric images unavailable in the baseline's one-way framework, we combine it with Shared CAM. Compared to using Shared CAM alone, Pixel Alignment further improves performance by \todo{8.7\%}, demonstrating its ability to refine initial knowledge activation maps into more regionally complete object segments.
\textbf{(5) Shared CAM with Denoising Distillation}: This combination achieves an average performance increase of \todo{7.5\%} over using either component independently, suggesting that refined exocentric information better bridges the domain gap while shared knowledge extraction enhances final affordance localization.
\textbf{(6) Complete Model}: Our full model achieves 1.088, 0.443, and 1.322 on KLD, SIM, and NSS metrics respectively on the seen split, demonstrating the synergistic effectiveness of all proposed components.

\begin{figure}
    \centering
    \includegraphics[width=1\linewidth]{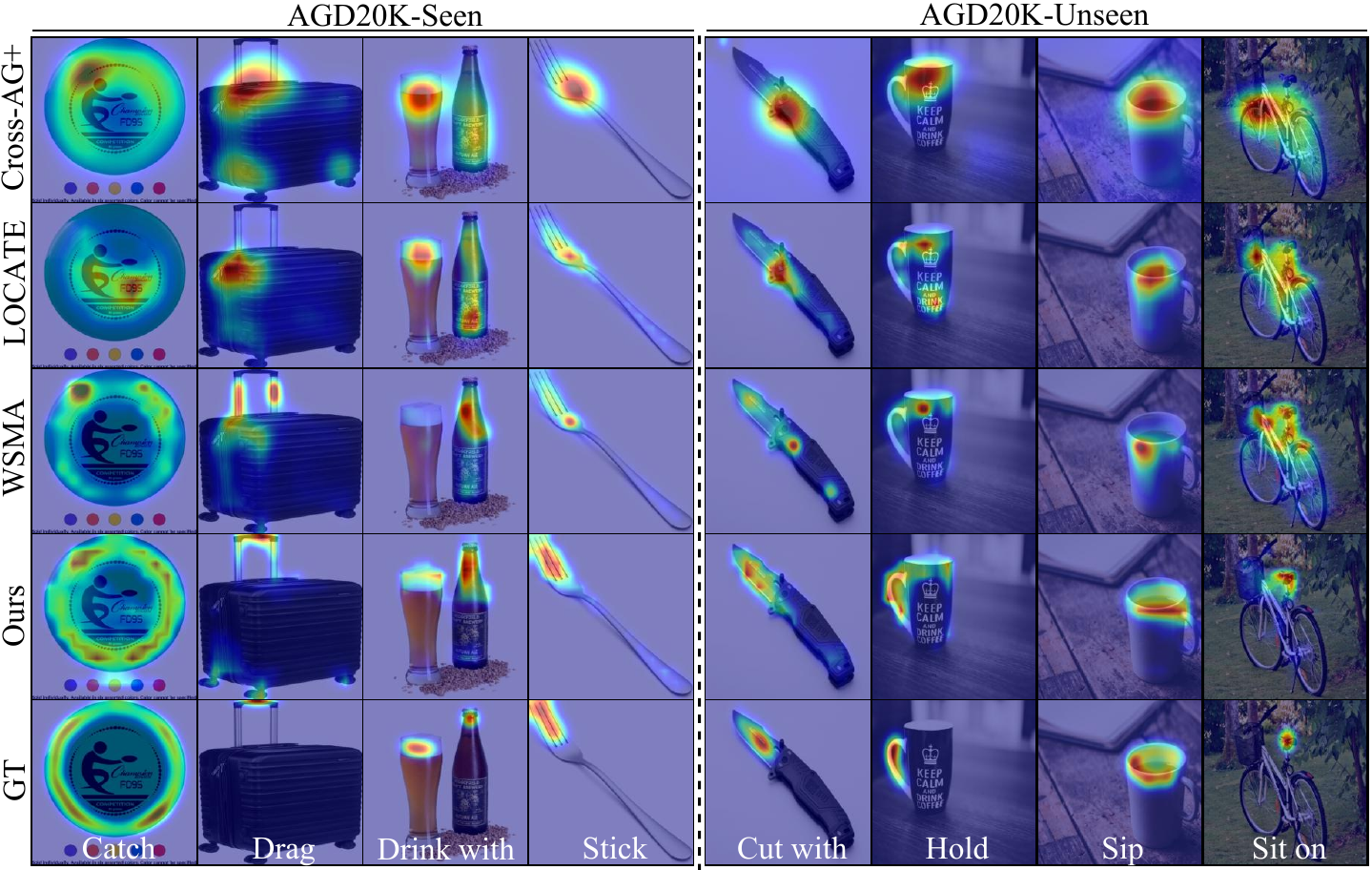}
    \vspace{-4mm}
    \caption{Visualization results compared with existing methods.}
    \label{fig:5}
    \vspace{-4mm}
\end{figure}

\subsection{Visualization}
As shown in Figure~\ref{fig:5}, we present additional qualitative affordance grounding results. Compared to Cross-View-AG+~\cite{luo2024grounded}, LOCATE~\cite{li2023locate}, and WSMA~\cite{xu2024weakly}, our approach significantly outperforms them in both localization accuracy and completeness. Notably, for occlusion-prone affordances like ``sit on" and ``catch", our method achieves precise localization, while previous methods struggled due to occlusion. Moreover, our pixel-level decoding ensures that localized regions are precise, unlike the vague and broad results of prior methods. Additionally, our closed-loop knowledge transfer enables significantly better performance on challenging unseen splits compared to feature-based one-way approaches. 
\section{Conclusion}
\label{sec:conclusion}
This paper proposes a closed-loop knowledge transfer framework for weak affordance localization. By incorporating shared knowledge extraction, pixel-level alignment, and positive knowledge feedback with denoising distillation, our model achieves consistent and significant improvements across all benchmarks. 
\clearpage
\noindent\textbf{Acknowledgment:} This work is supported by the National Natural Science Foundation of China (No.62206174). 
{
    \small
    \bibliographystyle{ieeenat_fullname}
    \bibliography{main}
}
\clearpage

\section{Challenges of the Current One-Way Transfer Framework}
\label{sec:a}
\begin{figure}[h]
    \centering
    \includegraphics[width=1\linewidth]{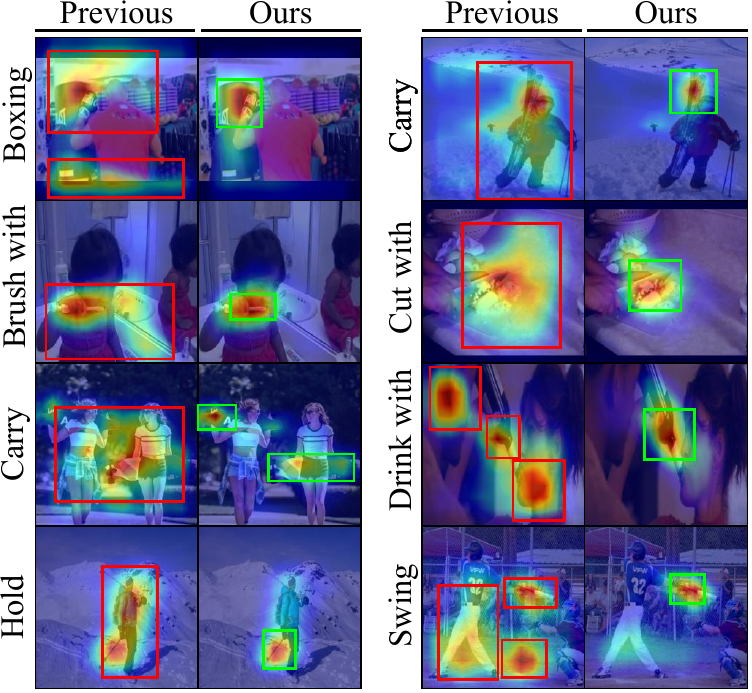}
    \caption{Comparison of knowledge activation on exocentric images between our LoopTrans and the previous one-way transfer framework.}
    \label{fig:a1}
\end{figure}

As discussed in the main text, two essential yet challenging aspects of weakly supervised affordance grounding are: (1) extracting affordance knowledge from exocentric interactions and (2) transferring it to egocentric localization.
\noindent \textbf{Challenge of Vague and Broad Activation.}
For knowledge extraction, previous one-way pipelines~\cite{luo2022learning, luo2024grounded, li2023locate, xu2024weakly} often struggle to generate accurate exocentric activation maps due to interference from complex and diverse interaction scenarios, such as background clutter and the presence of human bodies. This makes it difficult to capture the correct affordance regions. As shown in Figure~\ref{fig:a1}, the previous method~\cite{li2023locate} produces coarse or even erroneous knowledge activation maps. In contrast, our approach leverages a shared CAM mechanism that integrates object-centric and interaction-centric knowledge learning, enabling precise activation focused on affordance-relevant object regions.

\begin{figure}[h]
    \centering
    \includegraphics[width=1\linewidth]{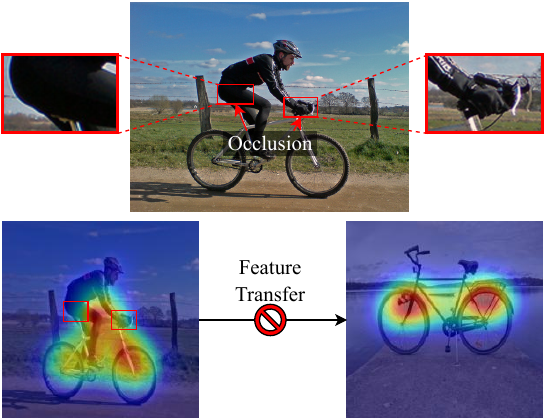}
    \caption{Ineffective knowledge feature extraction and transfer caused by occlusion in one-way transfer frameworks.}
    \vspace{-5mm}
    \label{fig:a2}
\end{figure}

\begin{table*}[h]
    \centering
    \setlength\tabcolsep{2mm}
    \renewcommand\arraystretch{0.9}
    \vspace{2mm}
    \begin{tabular}{clccccccccccc}
    \toprule
    & \multirow{2.5}{*}{\textbf{Method}} & \multicolumn{3}{c@{\hskip 0.1in}}{\textbf{Big}} && \multicolumn{3}{c@{\hskip 0.1in}}{\textbf{Middle}} && \multicolumn{3}{c@{\hskip 0.1in}}{\textbf{Small}} \\
    \cmidrule(r){3-5} \cmidrule(r){7-9} \cmidrule(r){11-13}
    &   & KLD$\downarrow$ & SIM$\uparrow$ & NSS$\uparrow$ && KLD$\downarrow$ & SIM$\uparrow$ & NSS$\uparrow$ && KLD$\downarrow$ & SIM$\uparrow$ & NSS$\uparrow$ \\
    \midrule
    \multirow{8}{*}{\rotatebox{90}{\textbf{Seen}}} 
    & EIL~\cite{mai2020eil} & 1.047& 0.461& 0.389&& 1.794& 0.284& 0.710&& 3.057& 0.123& 0.231\\
    & SPA~\cite{pan2021spa} & 5.745& 0.317& 0.222&& 4.990& 0.228& 0.440&& 6.076& 0.118& 0.297\\
    & TS-CAM~\cite{gao2021ts} & 1.039& 0.424& 0.166&& 1.814& 0.248& 0.401&& 2.652& 0.132& 0.352\\
    & Hotspots~\cite{nagarajan2019grounded} & 0.986& 0.448& 0.408&& 1.738& 0.265& 0.672&& 2.587& 0.149& 0.683\\
    & Cross-View-AG~\cite{luo2022learning} & 0.766& 0.533& 0.652&& 1.485& 0.322& 1.040&& 2.373& 0.175& 0.927\\
    & Cross-View-AG+~\cite{luo2024grounded} & 0.787& 0.521& 0.660&& 1.481& 0.314& 1.089&& 2.381& 0.167& 0.959\\
    & LOCATE~\cite{li2023locate} & 0.676& 0.580& 0.706&& 1.178& 0.390& 1.316&& 2.029& 0.216& 1.349\\
    & \textbf{Ours} & \textbf{0.661} & \textbf{0.607} & \textbf{0.737} && \textbf{1.069} & \textbf{0.435} & \textbf{1.411} && \textbf{1.657} & \textbf{0.276} & \textbf{1.774}\\
    \midrule
    
    \multirow{8}{*}{\rotatebox{90}{\textbf{Unseen}}} 
    & EIL~\cite{mai2020eil} & 1.199& 0.393& 0.271&& 1.906& 0.246& 0.482&& 3.082& 0.113& 0.116\\
    & SPA~\cite{pan2021spa} & 8.299& 0.259& 0.254&& 6.938& 0.186& 0.333&& 7.784& 0.095& 0.144\\
    & TS-CAM~\cite{gao2021ts} & 1.238& 0.351& 0.072&& 1.970& 0.208& 0.236&& 2.766& 0.113& 0.124\\
    & Hotspots~\cite{nagarajan2019grounded} & 1.015& 0.425& 0.548&& 1.872& 0.242& 0.605&& 2.693& 0.134& 0.544\\
    & Cross-View-AG~\cite{luo2022learning} & 0.884& 0.500& 0.728&& 1.595& 0.303& 0.945&& 2.558& 0.147& 0.692\\
    & Cross-View-AG+~\cite{luo2024grounded} & 0.867& 0.485& 0.776&& 1.658& 0.279& 0.988&& 2.630& 0.133& 0.754\\
    & LOCATE~\cite{luo2024grounded} & 0.571& 0.629& 0.956&& 1.302& 0.373& 1.257&& 2.223& 0.189& 1.071\\
    & \textbf{Ours} & \textbf{0.568} & \textbf{0.619} & \textbf{1.021} && \textbf{1.140} & \textbf{0.417} & \textbf{1.422} && \textbf{1.965} & \textbf{0.223} & \textbf{1.355}\\
    \bottomrule
    \end{tabular}
    \caption{Comparison results on AGD20K with different affordance region scales.}
    \label{tab:a1}
\end{table*}

\noindent \textbf{Invalid Transfer Caused by Occlusion Challenge.}
For the exocentric-to-egocentric transfer, one-way pipelines aim to achieve feature-based transfer by leveraging interaction knowledge in the exocentric view and aligning features between the exocentric and egocentric perspectives. However, in most scenarios involving actions like riding or holding, the interaction areas on the objects are often occluded by the human body due to the nature of these actions.  As illustrated in Figure~\ref{fig:a2}, even if the vague activation regions indicated by previous methods are roughly correct, occlusion by the human body prevents the accurate extraction of affordance-related object features. This renders the knowledge (features) transferred by one-way pipelines ineffective. In contrast, our approach jointly leverages shared exocentric interaction knowledge and egocentric object knowledge, rather than relying on explicit feature transfer. This allows interaction knowledge to be directly activated on egocentric images, fundamentally addressing the issue of ineffective feature-based transfer caused by occlusion.

\section{Details of Datasets and Metrics}
\label{sec:b}
\noindent \textbf{Image Datasets.}
AGD20K~\cite{luo2022learning} is a large-scale dataset specifically designed for affordance grounding, comprising 20,061 exocentric images and 3,755 egocentric images annotated with 36 distinct affordance categories. This dataset facilitates evaluations in both Seen and Unseen settings, allowing us to assess the model's ability to generalize across different object categories. HICO-IIF~\cite{xu2024weakly}, a composite dataset derived from HICO-DET~\cite{chao2018learning} and IIT-AFF~\cite{nguyen2017object}, includes exocentric images from HICO-DET and egocentric images from IIT-AFF, covering ten affordance classes and seven object categories. This combination enables performance evaluation even with a relatively limited dataset size, consisting of 4,383 training images and 1,498 test images.

\noindent \textbf{Video Datasets.}
OPRA~\cite{fang2018demo2vec} comprises approximately 16,000 product review videos sourced from YouTube, showcasing interactions with household appliances. Each video is paired with a static product image, an action label, and an affordance heatmap that highlights relevant interaction regions. EPIC-Kitchens~\cite{damen2018scaling} features unscripted egocentric videos depicting various kitchen activities, annotated with action and object labels. Target images are selected from specific frames, and crowd-sourced annotations provide ground-truth heatmaps for relevant affordance regions. Together, these datasets enhance our ability to learn affordance grounding from both images and videos of human-object interactions.

\noindent \textbf{Details of Metrics}
\begin{itemize}
    \item \textbf{Kullback-Leibler Divergence (KLD):} The Kullback-Leibler Divergence quantifies the divergence between two probability distributions. The formula for KLD is:

    \begin{equation}
    \text{KLD}(P, Q) = \sum_i Q_i \log \left(\frac{Q_i}{P_i}\right).
    \end{equation}

    Here, \( P \in \mathbb{R}^{HW} \) is the predicted heatmap, and \( Q \in \mathbb{R}^{HW} \) is the true distribution. \( H \) and \( W \) represent the height and width of the distributions, respectively.

    \item \textbf{Similarity (SIM):} The Similarity metric assesses the degree of overlap between the predicted and true distributions. The formula for SIM is:

    \begin{equation}
    \text{SIM}(P, Q) = \sum_i \min(P_i, Q_i).
    \end{equation}

    Here, \( P \in \mathbb{R}^{HW} \) is the predicted distribution, and \( Q \in \mathbb{R}^{HW} \) is the true distribution.

    \item \textbf{Normalized Scanpath Saliency (NSS):} Normalized Scanpath Saliency evaluates how well the predicted heatmap aligns with the ground truth binary map. Specifically, we first normalize the predicted distribution and then calculate the NSS:

    \begin{equation}
    \begin{aligned}
        &\bar{P} = \frac{P - \mu(P)}{\sigma(P)},\\
        &\text{NSS}(\bar{P}, M) = \frac{1}{HW}\sum_i \bar{P}_i \times M_i.
    \end{aligned}
    \end{equation}

    Here, \( P \in \mathbb{R}^{HW} \) is the predicted heatmap, \( M \in \{0, 1\}^{HW} \) is the ground truth binary map, and \(\mu(P)\) and \(\sigma(P)\) are the mean and standard deviation of \( P \), respectively.
\end{itemize}

\noindent \textbf{Details of Image-to-Video Generalization Setting.}
Since the affordance categories in OPRA and EPIC differ from those in AGD20K, models trained on AGD20K cannot be directly tested on these video datasets. However, we observe that due to the richness of affordance categories in AGD20K, most affordances in other datasets have similar counterparts in AGD20K. Thus, we align the affordance categories by mapping each category in the video dataset to its most relevant counterpart in the image dataset for testing.

\section{More Experimental Results}
\label{sec:c}
\noindent \textbf{Comparison on Different Scales.}
Following~\cite{li2023locate,luo2022learning}, we divide the test set into three subsets—large, medium, and small—based on the size of the affordance object regions. The subsets are defined as follows: affordance regions occupying more than 10\% of the image area are categorized as large, those between 3\% and 10\% as medium, and those smaller than 3\% as small. As shown in Table~\ref{tab:a1}, we conduct experiments on AGD20K~\cite{luo2022learning} dataset to assess the robustness of our approach across different affordance region scales. The results demonstrate that our method consistently outperforms previous approaches~\cite{luo2022learning,luo2024grounded,li2023locate} across all size categories.

\begin{table}[h]
    \centering
    \small
    \setlength\tabcolsep{1.5mm}
    \renewcommand\arraystretch{1}
    \begin{tabular}{ccccccc}
        \toprule
        \multirow{2.5}{*}{\textbf{Exo:Ego}} & \multicolumn{3}{c@{\hskip 0.1in}}{\textbf{AGD20K-Seen}} & \multicolumn{3}{c@{\hskip 0.1in}}{\textbf{AGD20K-Unseen}}\\
        \cmidrule(r){2-4} \cmidrule(r){5-7} 
            & KLD$\downarrow$ & SIM$\uparrow$ & NSS$\uparrow$ & KLD$\downarrow$ & SIM$\uparrow$ & NSS$\uparrow$ \\
        \midrule
        1:1 & 1.077 & 0.449 & 1.335 & 1.250 & \textbf{0.411} & 1.312 \\
        2:1 & 1.097 & 0.445 & 1.321 & 1.265 & 0.405 & 1.306 \\
        \rowcolor{pink!40}
        3:1 & 1.088 & 0.445 & 1.322 & \textbf{1.247} & 0.403 & \textbf{1.315} \\
        4:1 & \textbf{1.068} & \textbf{0.452} & \textbf{1.341} & 1.285 & 0.396 & 1.301 \\
        5:1 & 1.103 & 0.443 & 1.309 & 1.265 & 0.402 & 1.307 \\
        \bottomrule
    \end{tabular}
    \vspace{-2mm}
    \caption{Ablation study on the ratio of exocentric images to ego images}
    \label{}
    \vspace{-2mm}
\end{table}
\noindent \textbf{Ablation on Image Radio.}
The ablation study evaluates the effect of varying exocentric-to-egocentric sample ratios (1:1 to 5:1) on AGD20K-Seen and AGD20K-Unseen datasets. On AGD20K-Seen, the 4:1 ratio achieves the best performance across all metrics (KLD: 1.068, SIM: 0.452, NSS: 1.341), indicating that emphasizing exocentric data enhances affordance feature extraction while maintaining egocentric localization. However, performance declines at 5:1, suggesting diminishing returns with excessive exocentric emphasis. For AGD20K-Unseen, the 3:1 ratio performs best (NSS: 1.315, KLD: 1.247, SIM: 0.403), demonstrating robust generalization to unseen affordance objects. Further increases in the ratio reduce SIM and NSS, underscoring the need for sufficient egocentric samples to support generalization. For a fair comparison with prior methods~\cite{li2023locate,luo2022learning}, we adopt the 3:1 ratio as the final setting.

\begin{figure}
    \centering
    \includegraphics[width=1\linewidth]{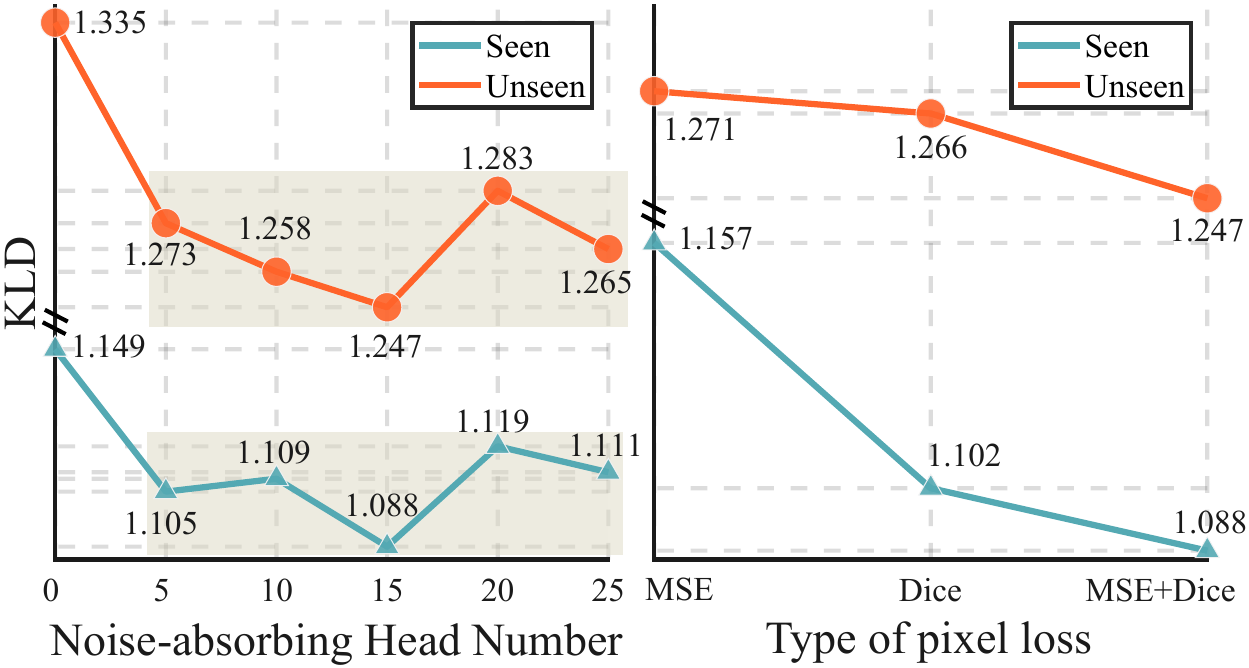}
    \vspace{-3mm}
    \caption{Ablation study on the number of noise-absorbing heads (left) and type of pixel decoder loss (right). }
    \label{fig:4}
    \vspace{-5mm}
\end{figure}

\noindent \textbf{Ablation on the Number of Noise-absorbing Heads.}
As shown in Figure~\ref{fig:4} (left), we conduct experiments to assess the impact of the number of noise-absorbing heads. Setting the number of noise-absorbing heads to zero—indicating a purely distilled, non-denoising approach—leads to significantly reduced performance, as noisy features in exocentric images disrupt knowledge extraction and hinder exocentric-to-egocentric transfer. Increasing the number of heads improves performance, particularly on the seen split, with optimal results at 15.

\noindent \textbf{Ablation on the Type of Pixel Decoder Loss.}
 As shown in Figure~\ref{fig:4} (right), we conduct experiments to assess the impact of pixel decoder loss type. The results shows that combining MSE and Dice losses markedly outperforms either loss alone, especially on the unseen split, where the pixel-level Dice loss enhances cross-category localization alignment.

\section{More Visualizations}
\label{sec:d}
\begin{figure*}
    \centering
    \vspace{1mm}
    \includegraphics[width=1\linewidth]{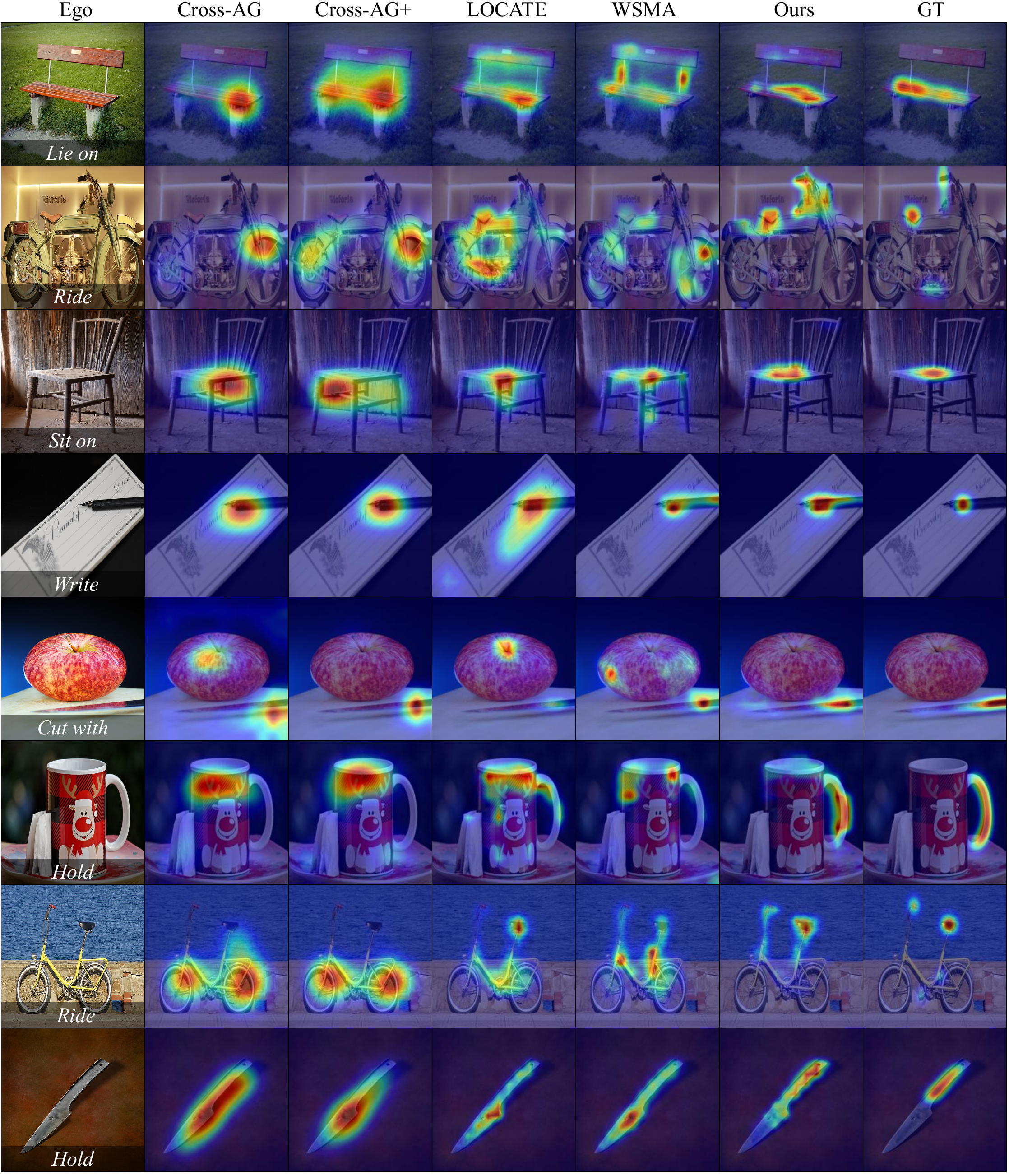}
    \vspace{1mm}
    \caption{More visualization for affordance grounding results on egocentric images.}
    \label{fig:a3}
\end{figure*}

\begin{figure*}
    \centering
    \vspace{1mm}
    \includegraphics[width=1\linewidth]{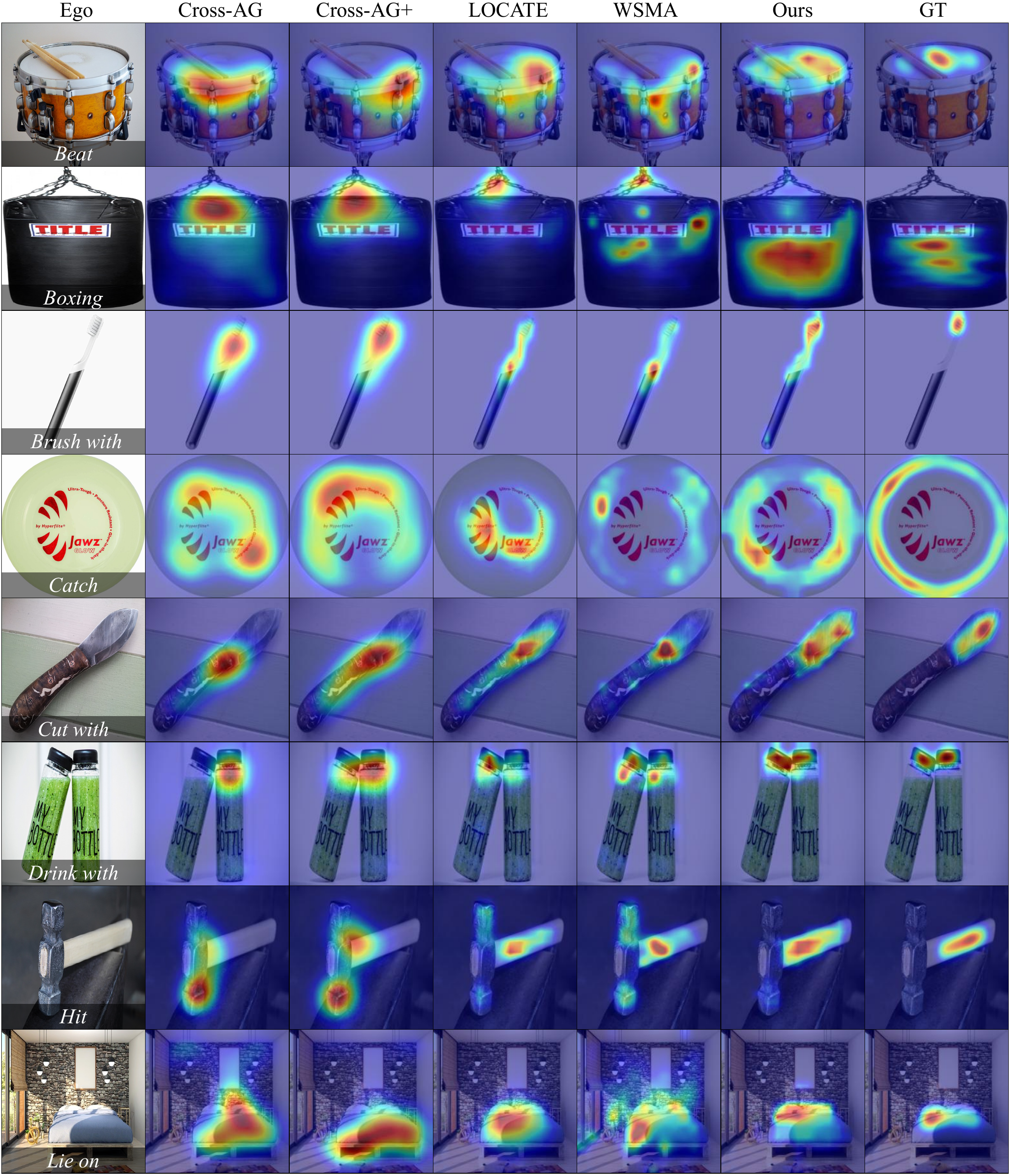}
    \vspace{1mm}
    \caption{More visualization for affordance grounding results on egocentric images.}
    \label{fig:a4}
\end{figure*}

As illustrated in Figure~\ref{fig:a3} and Figure~\ref{fig:a4}, we present visualization results across multiple affordance categories. Our method consistently surpasses all existing approaches~\cite{luo2022learning,luo2024grounded,li2023locate,xu2024weakly} across all categories, demonstrating notable advantages in scenarios where affordance regions are occluded by human interactions. For example, in the \textit{hold} and \textit{ride} categories, our approach effectively localizes occluded regions such as handles and saddle areas in egocentric images—achievements that remain unattainable by prior methods.

\clearpage

\end{document}